\documentclass{article}

\usepackage{microtype}
\usepackage{graphicx}
\usepackage{caption}
\usepackage{subcaption}
\usepackage{booktabs}

\usepackage{amsmath,amssymb,amsfonts,nicefrac}
\usepackage{amstext}
\usepackage{pgfplots}
\pgfplotsset{compat = 1.3}
\usepackage{multirow}
\usepackage{makecell}
\usepackage{booktabs}
\usepackage{listings}
\usepackage[ruled,vlined]{algorithm2e}

\usepackage{eucal}
\usepackage{ mathrsfs }

\usepackage{arxiv}
\renewcommand{\headeright}{}
\renewcommand{\undertitle}{Technical Report}

\def\x{{\mathbf x}}
\def\y{{\mathbf y}}
\def\yhat{\hat \y}

\def\tco{\texttt{train-clean-100}}
\def\tct{\texttt{train-clean-360}}
\def\tof{\texttt{train-other-500}}
\def\devclean{\texttt{dev-clean}}
\def\devother{\texttt{dev-other}}
\def\testclean{\texttt{test-clean}}
\def\testother{\texttt{test-other}}

\newcommand{\librispeech}{\textsc{LibriSpeech}}
\renewcommand{\paragraph}[1]{\textbf{#1:}}

\usepackage{soul}

\title{Population Based Training for Data Augmentation and Regularization in Speech Recognition}
\author{Daniel Haziza \\
Facebook AI Research, Paris, France \\
\texttt{dhaziza@fb.com}
\And
J\'{e}r\'{e}my Rapin \\
Facebook AI Research, Paris, France \\
\texttt{jrapin@fb.com} \And
Gabriel Synnaeve \\
Facebook AI Research, New York, USA \\
\texttt{gab@fb.com}}

\begin{document}

\maketitle
 
\begin{abstract}
Varying data augmentation policies and regularization over the course of optimization has led to performance improvements over using fixed values.
We show that population based training is a useful tool to continuously search those hyperparameters, within a fixed budget.
This greatly simplifies the experimental burden and computational cost of finding such optimal schedules.
We experiment in speech recognition by optimizing SpecAugment \cite{park2019specaug} this way, as well as dropout. It compares favorably to a baseline that does not change those hyperparameters over the course of training, with an 8\% relative WER improvement. We obtain 5.18\% word error rate on LibriSpeech's \texttt{test-other}.
\end{abstract}

\section{Introduction and Related Work}
\label{sec:introduction}
Deep neural networks tend to benefit from training with as much data as available. As high-quality in-domain data is usually scarce, recent advances in the automatic discovery of data augmentation policies \cite{ho2019population,lim2019fast,cubuk2019randaugment} may extend the applicability of neural networks to smaller datasets, or simply their ability to generalize. However, finding good data augmentation policies remains computationally expensive. An obvious reason is the size of the search space, another is that the other hyperparameters are conditioned on the (stronger) data augmentation policies. Because of that computational cost, most data augmentation research is done either on proxy datasets or on smaller models (proxy tasks). Moreover, some experimental findings suggest that having a schedule over the data augmentation policy may be helpful for being able to train better (or, at all) with strong policies. Having a schedule -- or a curriculum \cite{bengio2009curriculum} -- is common for other hyperparameters like the learning rate \cite{bottou2012stochastic,loshchilov2016sgdr} or task difficulty \cite{sukhbaatar2017intrinsic} but rarely used for data augmentation as the search space for a schedule is even bigger than for a fixed policy.

The absolute state of the art on \librispeech \cite{panayotov2015librispeech} (English read speech) \cite{panayotov2015librispeech} is obtained through semi-supervision \cite{synnaeve2019e2e}, while the best results on \librispeech~only are held by hybrid systems \cite{wang2019transformerbased,luscher2019transformers}, followed closely by end-to-end systems \cite{synnaeve2019e2e,park2019specaug}. In speech, and audio more generally, data augmentation has been used for decades, from speed-related perturbations \cite{ko2015audio}, vocal tract length perturbation \cite{jaitly2013vocal}, time-frequency additions/subtractions \cite{park2019specaug,Tth2018API}, to applying GANs to adapt to new domains \cite{hosseini2018multi} or to generalize to new speakers \cite{kaneko2019cyclegan}. In particular, SpecAugment \cite{park2019specaug}, that we use and study in this paper, was a very impactful addition to ASR data augmentation.

Population based training (PBT) \cite{Jaderberg_17_PopulationBasedTraining} is often used in reinforcement learning, in part due to the dynamic nature of training policies, and sometimes more broadly in hyperparameter search, used similarly to (partially) iterative hyperparameter search methods \cite{li2017hyperband}. 
Population based augmentation (PBA) \cite{ho2019population} proposes to leverage PBT as a tunable parallel and iterative mechanism to search for good data augmentation policies. However, the settings in which it was tested are still limited in scale (CIFAR-10/100 and SVHN). 

AutoAugment \cite{cubuk2019autoaugment} proposed to use reinforcement learning to learn series of transformation to apply to input data, to discover data augmentation policies. It is on the high side of computationally expensive data augmentation policies search. Fast AutoAugment \cite{lim2019fast} speeds up this search process through density matching of the data and was applied to ImageNet. RandAugment \cite{cubuk2019randaugment} also tackles scaling up the search for data augmentation policies, but it strongly restricts the search space and searches for the policy online (instead of as a separate step). Population based augmentation \cite{ho2019population} jointly trains the final model while continuously searching for better data augmentation policies. It is the closest to our work, with the difference that we do not limit ourselves to data augmentation, and apply PBT at a larger scale.

Our contributions are the following:
    We show experimentally that PBT can train bigger models than in previous PBA studies, to state-of-the-art results on LibriSpeech
    surpassing the previous comparable model by 0.4\% WER on \testother{} after decoding.
    By tuning data augmentation policies and regularization parameters jointly and with enough diversity in the population, we show that exploration of many hyperparameters yields schedules (changing values over the course of training).

\section{Method}
\label{sec:method}
The idea behind population based training \cite{Jaderberg_17_PopulationBasedTraining} is quite simple: train multiple models at the same time and slightly mutate their hyperparameters randomly over time. Models which perform significantly poorer than others are stopped, and the freed computational resources are used to train copies of the better ones. These copies will then diverge from their parent due to mutations accumulating over time. Repeating these steps will selects best performing hyperparameters, while also making the training process robust against diverging models.

\subsection{Hyperparameters tuning with PBT}

We train a model for a fixed number of iterations or epochs between two checkpoints and mutations, and we call this a \emph{training step}. For a given checkpoint, we define its generation as the number of training steps that separate it from the random initialization.
The process is described in Algorithm~\ref{listing:worker}.

\lstset{basicstyle=\ttfamily}

\begin{lstlisting}[language=python,caption={Training node code},label={listing:worker}]
while True:
    ckpt, hparams = find_parent_to_train()
    hparams = mutate(hparams)
    new_ckpt = train(ckpt, hparams)
    loss = evaluate(new_ckpt)
    report_result(new_ckpt, loss)
\end{lstlisting}

\paragraph{Population management} Every time we finish a training step, we need to decide if the model can continue training, or should be stopped. We use initiator based evolution introduced in \cite{li2019generalized}. Every model is compared to a randomly sampled opponent model in the population: if the loss of the opponent model is significantly better, the initial model is killed, and its resources are used to train a fork of the opponent model. We provide more details about this procedure in the appendix.

\paragraph{Hyperparameters mutation} Before each training step, we change the hyperparameter values randomly. For the model to be stable, we want those parameters to be close to the parent's, so their values are only slightly perturbed. In practice, we randomly add or subtract a constant, while ensuring that the value remains within a predefined range.

\subsection{Acoustic Models}
\label{sec:am}
Our transformer-based acoustic models have a small front-end which takes 80 Mel filterbanks as input: $6$ layers of 1D convolutions each of kernel width $3$ and respective input and output sizes $(80,2048)$, $(1024,2048)$, $(1024,2048)$, $(1024,2048)$, $(1024,2048)$, $(1024,1536)$, respectively. Each convolution is followed by a GLU activation function \cite{dauphin2017gcnn} and every other convolution strides by 2. The output of the front-end is thus strided by $8$ frames (80 milliseconds). After the front-end, there are $36$ Transformer blocks, and each block has $4$ self-attention heads with dimension $768$, followed by a feedforward network (FFN), one hidden layer of dimension $3072$, and ReLU non-linearity. Precisely, given a sequence of $T$ vectors of dimension $d$, the input is represented by the matrix $\mathbf{H^0} \in \mathbb{R}^{d \times T}$, following exactly \cite{vaswani2017attention}, we have: 
\begin{align*}
\mathbf{Z}^{i} & = \textsc{Norm}(\textsc{SelfAttention}(\mathbf{H}^{i-1}) + \mathbf{H}^{i-1}), \\
\mathbf{H}^{i} & = \textsc{Norm}(\textsc{FFN}(\mathbf{Z}^i) + \mathbf{Z}^i),
\end{align*}
where $\mathbf{Z}$ is the output of the self-attention layer, with a skip connection, and $\mathbf{H}$ is the output of the FFN layer, with a skip connection. As is standard: our norm is layer norm, and self-attention is defined as:
\begin{align*}
\mathbf{S}^r = \textsc{SoftMax}(\frac{1}{\sqrt{d}} (\mathbf{W}_K\mathbf{H})^\top (\mathbf{W}_Q\mathbf{H})_t^{r-1})\mathbf{W}_V\mathbf{H}.
\end{align*}
As we use Seq2Seq models, we have an additional decoder, which is a stack of 6 Transformers, with encoding dimension 256, and 4 attention heads. The probability distribution of the transcription is factorized as:
\begin{equation}
p(y_1, ..., y_n) = \prod_{i=1}^n p(y_i \ | \ y_0, ..., y_{i-1}, \mathbf{H}^{L_e}),
\end{equation}
where $y_0$ is a special symbol indicating the beginning of the transcription. For all layers (encoder and decoder): we use dropout on the self-attention, and we also use layer drop \cite{fan2019reducing}, dropping entire layers at the FFN level.

We use SpecAugment~\cite{park2019specaug}, in our baseline we use exactly the LD policy: 2 masks of up to 27 frequency bands (for 80 filterbanks), and 2 masks of up to 1 second in time, all masks are sampled uniformly.
\section{Experiments}

\begin{table*}[t]
\begin{center}
\caption{Word error rates on \librispeech's development and test sets.}
\label{tab:libriWER}
\begin{small}
    \centering
    \begin{sc}
    \begin{tabular}{lccccccc}
    \toprule
       \multicolumn{2}{c}{AM} & \multicolumn{2}{c}{LM} & \multicolumn{2}{c}{Dev} & \multicolumn{2}{c}{Test} \\
    \cmidrule(lr){1-2} \cmidrule(lr){3-4} \cmidrule(lr){5-6} \cmidrule(lr){7-8}
        \multicolumn{1}{c}{type} & lexicon & type  & lexicon& clean & other & clean & other \\
    \midrule
    LAS \cite{park2019specaug} & 16k WP & - & - & & & 2.8 & 6.8 \\
    ~~~~ Decoding & 16k WP & RNN & 16k WP & & & 2.5 & 5.8 \\
    HMM/biLSTM \cite{luscher2019transformers} & 12k CDp & 4gram+LSTM & word & 2.2 & 5.1 & 2.6 & 5.5 \\
    ~~~~ + Transf. rescoring & 12k CDp & + Transf. & word & 1.9 & 4.5 & 2.3 & 5.0 \\
    Conv. Transf. \cite{wang2019transformerbased} & chenones & 4gram & word &  &  & 2.60 & 5.59 \\
    ~~~~ + Transf. rescoring & chenones & Transf. & word & & & 2.26 & 4.85 \\
    \midrule
    Conv. Transf. (270M) \cite{synnaeve2019e2e} & 10k WP & - & - & 2.56 & 6.65 & 3.05 & 7.01 \\
    ~~~~ Decoding & 10k WP & 6gram & 10k WP & 2.28 & 5.88 & 2.56 & 6.15 \\
    ~~~~ Decoding & 10k WP & GCNN & 10k WP & 2.11 & 5.25 & 2.30 & \textit{5.64} \\
    ~~~~ ~~~~ + Rescoring & 10k WP & GCNN + Transf. & word & 2.17 & 4.67 & 2.31 & 5.18 \\
    \midrule
    \multicolumn{1}{c}{ \textbf{PBT (this paper):} } & \\
    Conv. Transf. (296M) & 10k WP & - & - & 2.39 & 6.04 & 2.83 & 6.19 \\
    ~~~~ Decoding & & 6gram & 10k WP & 2.21 & 5.40 & 2.45 & 5.69 \\
    ~~~~ Decoding & & GCNN & 10k WP & 2.14 & 4.85 & 2.36 & \textit{5.18} \\
    \bottomrule
    \end{tabular}
    \end{sc}
\end{small}
\end{center}
\end{table*}

To validate the population based training approach, we use a state-of-the-art acoustic model that we train end-to-end with the open source toolkit wav2letter++ \cite{pratap2018wav2letter}. One model is trained on 64 Tesla V100 GPUs with a total batch size of $384$ ($6 \times 64$) on Librispeech's training datasets (\tco, \tct and \tof), without any additional data.

We define one (PBT) training step as 2200 updates -- between 2 and 3 epochs -- and let our PBT algorithm train 8 models in parallel for 3 days (about 160 steps). PBT being asynchronous, the number of models of each generation can vary: we observed between 6 and 12 models per generation, 8.3 on average.

\subsection{Technical details}
\paragraph{Stochastic data augmentation} In SpecAugment, \lstinline{tmask_n} and \lstinline{fmask_n} respectively control the number of maskings along the temporal and the frequency dimensions. Because both of these parameters typically take integer values from $1$ to $10$, an increment or decrement can have significant impact on the training. In order to define smaller mutations, we express these parameters as floats, and sample the number of masks for each mini-batch. For instance, if $fmask\_n = N + p$ with $p \in [0, 1]$, we mask $N$ frequency bands with probability $1 - p$, and $N + 1$ frequency bands with probability $p$.

\paragraph{Decoding} We produce some results with the single pass beam search decoder from wav2letter++ \cite{pratap2018wav2letter} with a 6gram word pieces model or with a convolutional language model with Gated Linear Units (named GCNN) \cite{dauphin2017gcnn}, similarly to \cite{synnaeve2019e2e}. The decoder maximizes $\log P_{AM}(\yhat | \x) + \alpha_{\mathrm{LMweight}} \log P_{LM}(\yhat) + \beta_{\mathrm{EOSscore}}|\yhat|$. We tune the two hyperparameters of the beam search (language model weight and end-of-sequence token score) on the validation (\devclean{} and \devother{}) sets and then decode with the best hyperparameters on the test sets. 

\begin{figure}
    \centering
    \includegraphics[width=0.6\linewidth]{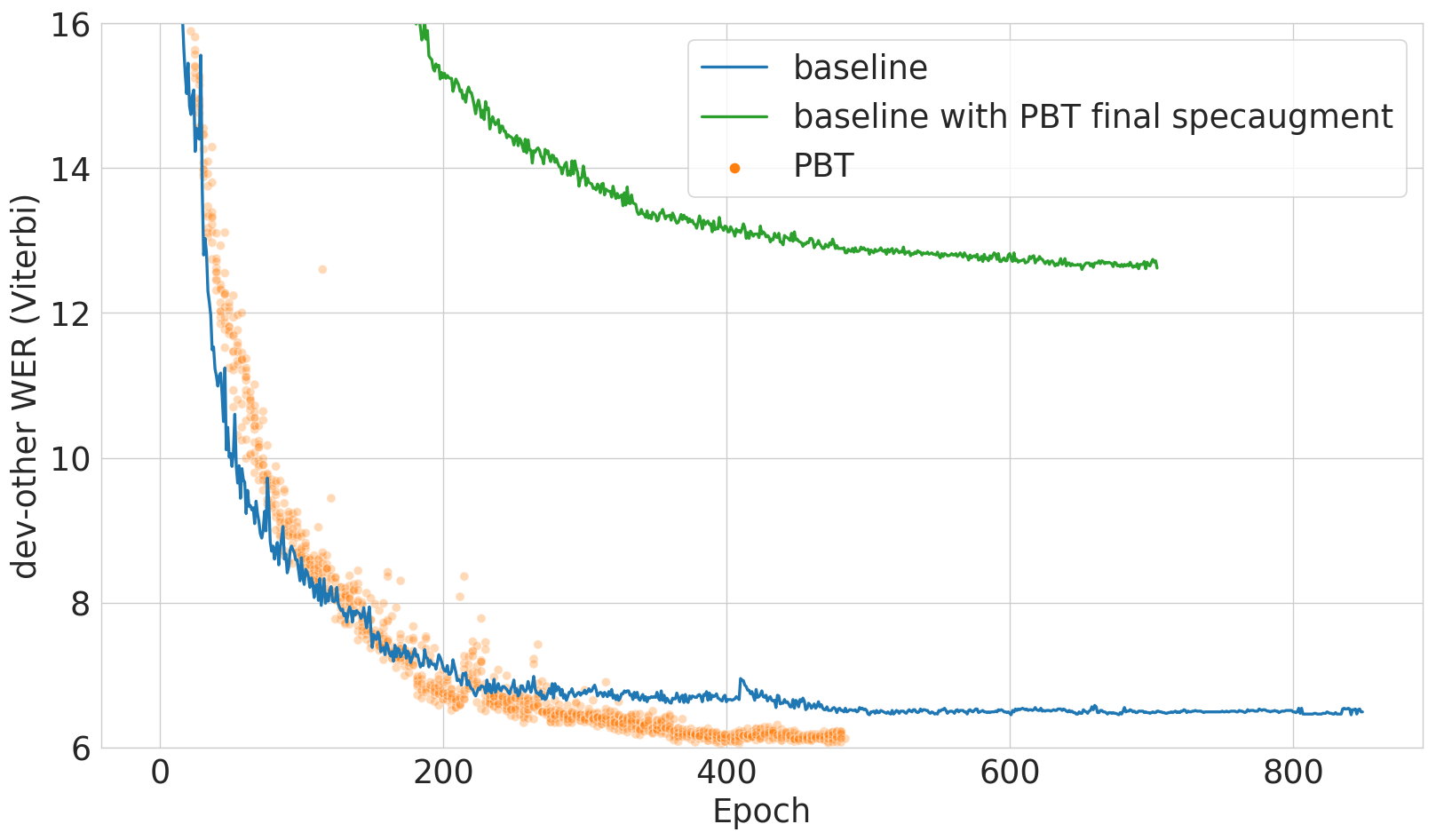}
    \subcaption{Word Error Rate on \librispeech{} \devother{} during training. The baseline is in blue - with fine-tuning starting after epoch 409, and each orange dot represents a checkpoint produced by the PBT run. In green, the same baseline has been repeated with the final values of PBT for data augmentation (Table~\ref{tab:mutation_settings})}
    \label{fig:w2l_devother_wer}
\end{figure}

\paragraph{Baseline} For comparison, we produce a baseline on the exact same acoustic model that we train with PBT. We use the parameters provided for SpecAugment in their original paper \cite{park2019specaug}, and optimal values from initial sweeps for the regularization (20\% of dropout and of layerdrop). All the other parameters are exactly the same as in the population based training run. The training curve of this baseline can be seen in blue in Figure~\ref{fig:w2l_devother_wer}. This model reaches 6.45\% dev-other WER (Viterbi) and 5.37\% dev-other WER with GCNN decoding. As it is outperformed by the similar model from \cite{synnaeve2019e2e} which gets 5.25\% with GCNN decoding, we directly compare to this (better) model in Table~\ref{tab:libriWER}.

\paragraph{Fitness measure and validation dataset}
Our training procedure selects, clones and culls individuals in the population based on a fitness metric that we measure after each training step. Given the relatively small size of Librispeech's validation dataset, and the small number of speakers (40) it contains, we have found that PBT can fairly easily overfit its distribution during longer trainings. Hence, we extracted a small part of Librispeech's \tof training dataset, and used it to evaluate our models trained with PBT. We ensured that this evaluation dataset contained exactly 20 male and 20 female speakers, each with the same number of sentences, and these speakers were excluded from the training data.

\paragraph{Data augmentation} Typically, researchers use constant values for data augmentation, and those values are usually high. However, strong data augmentation from the beginning of the training can cause the model to diverge. As a consequence, a common practice is to enable data augmentation only after the warmup period, which we do on our baseline. On our PBT run, we do not  rely on heuristics nor do we hard-code when it should be activated: data augmentation is enabled from the beginning, but with a very low magnitude. These data augmentation hyperparameters can then increase over the course of the training as better mutations are selected. The mutation scheme of these parameters are reported in the first part of Table~\ref{tab:mutation_settings}.

\paragraph{Regularization} We let PBT optimize 5 regularization parameters that determine values for dropout and layerdrop in the acoustic model and in the decoder. We reference their initial values and range, along with their mutation schemes in the second part of Table~\ref{tab:mutation_settings}.

\subsection{Results on ASR: LibriSpeech}

\begin{center}
\begin{table}[ht]
\caption{Initial, minimum value, maximum value and available mutations for augmentation parameters (first part) and regularization parameters (second part). For each parameter, we also report values for the \emph{baseline} (fixed) and for the best checkpoint on \devother{} averaged over the last 100 epochs of training (\emph{PBT}). Dropout and layerdrop related to transformer layers are prefixed with tr\_.}
\label{tab:mutation_settings}
\centering
\vskip 0.1in
\begin{small}
    \centering
\begin{tabular}{lcccccc}
\toprule
{parameter} & init. &   min &  max &        mutation & baseline & PBT \\
\midrule
\multicolumn{7}{l}{SpecAugment} \\
fmask\_f            &    7 &     7 &  120 &     $\pm 2.5$ or $5$ & 27.0 & $29.8$ \\
fmask\_n            &    1 &     1 &    8 &          $\pm 0.5$ & 2.0 & $1.7$ \\
tmask\_t            &   20 &    20 &  150 &       $\pm 2$ or $5$ & 100.0 & $54.1$ \\
tmask\_p            &  0.2 &   0.2 &    1 &  $\pm 0.05$ or $0.1$ & 1.0 & 0.95 \\
tmask\_n            &    1 &     1 &    8 &     $\pm 0.5$ or $1$ & 2.0 & 5.84 \\
\midrule
\multicolumn{7}{l}{Acoustic Model (encoder)} \\
dropout             &  0.2 &  0.01 &  0.8 &         $\pm 0.01$ & 0.2 & 0.23 \\
tr\_dropout          &  0.2 &  0.01 &  0.8 &         $\pm 0.01$ & 0.2 & 0.17 \\
tr\_layerdrop        &  0.2 &  0.01 &  0.8 &         $\pm 0.01$ & 0.2 & 0.27 \\
\midrule
\multicolumn{7}{l}{Acoustic Model (decoder)} \\
tr\_dropout   &  0.3 &  0.01 &  0.8 &         $\pm 0.01$ & 0.2 & 0.35 \\
tr\_layerdrop &  0.2 &  0.01 &  0.8 &         $\pm 0.01$ & 0.2 & 0.29 \\
\bottomrule  
\end{tabular}
\end{small}
\vskip -0.1in
\end{table}
\end{center}

\begin{figure}
    \centering
    \includegraphics[width=0.45\textwidth]{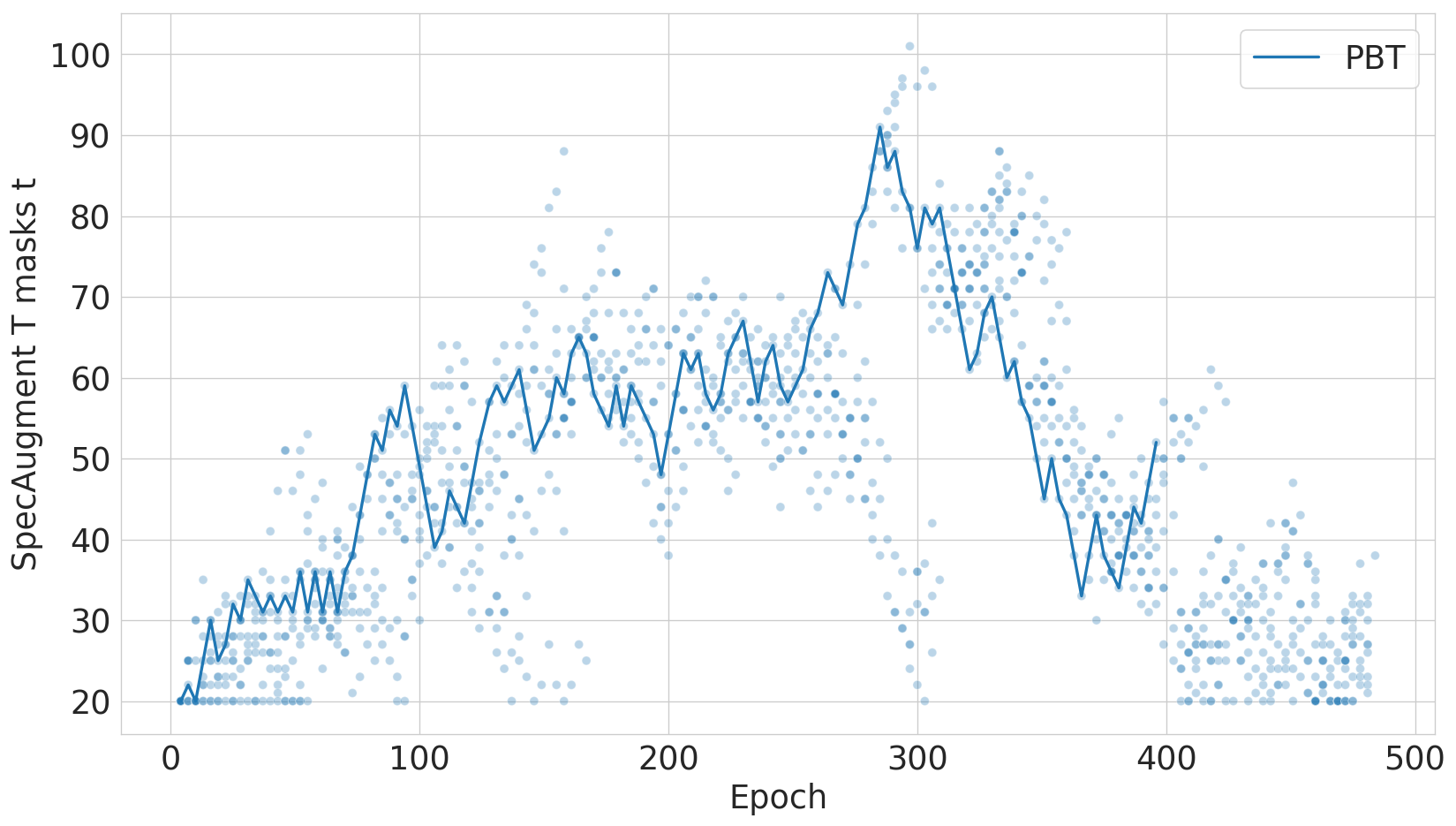}
    \includegraphics[width=0.45\textwidth]{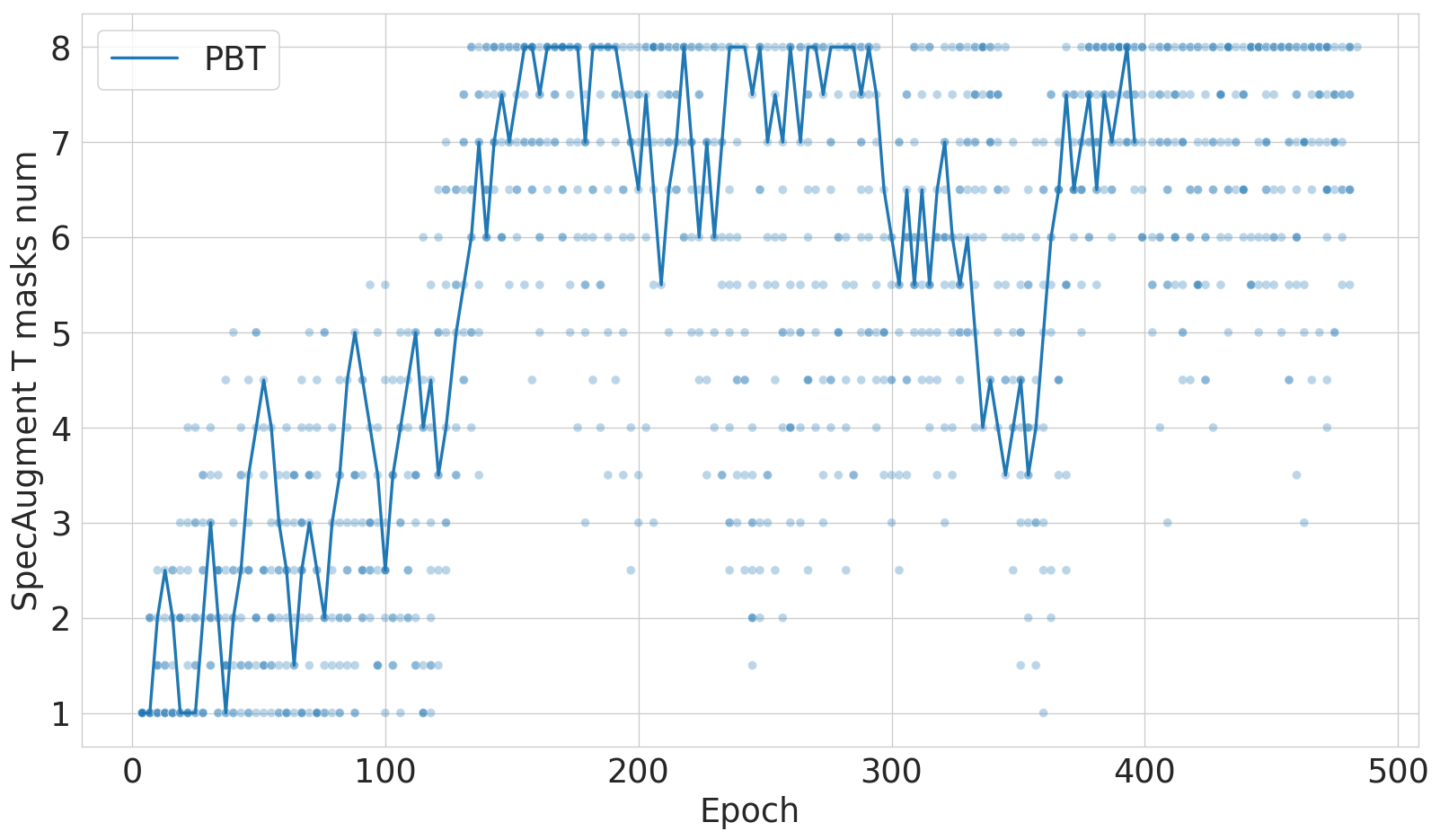}
    \includegraphics[width=0.45\textwidth]{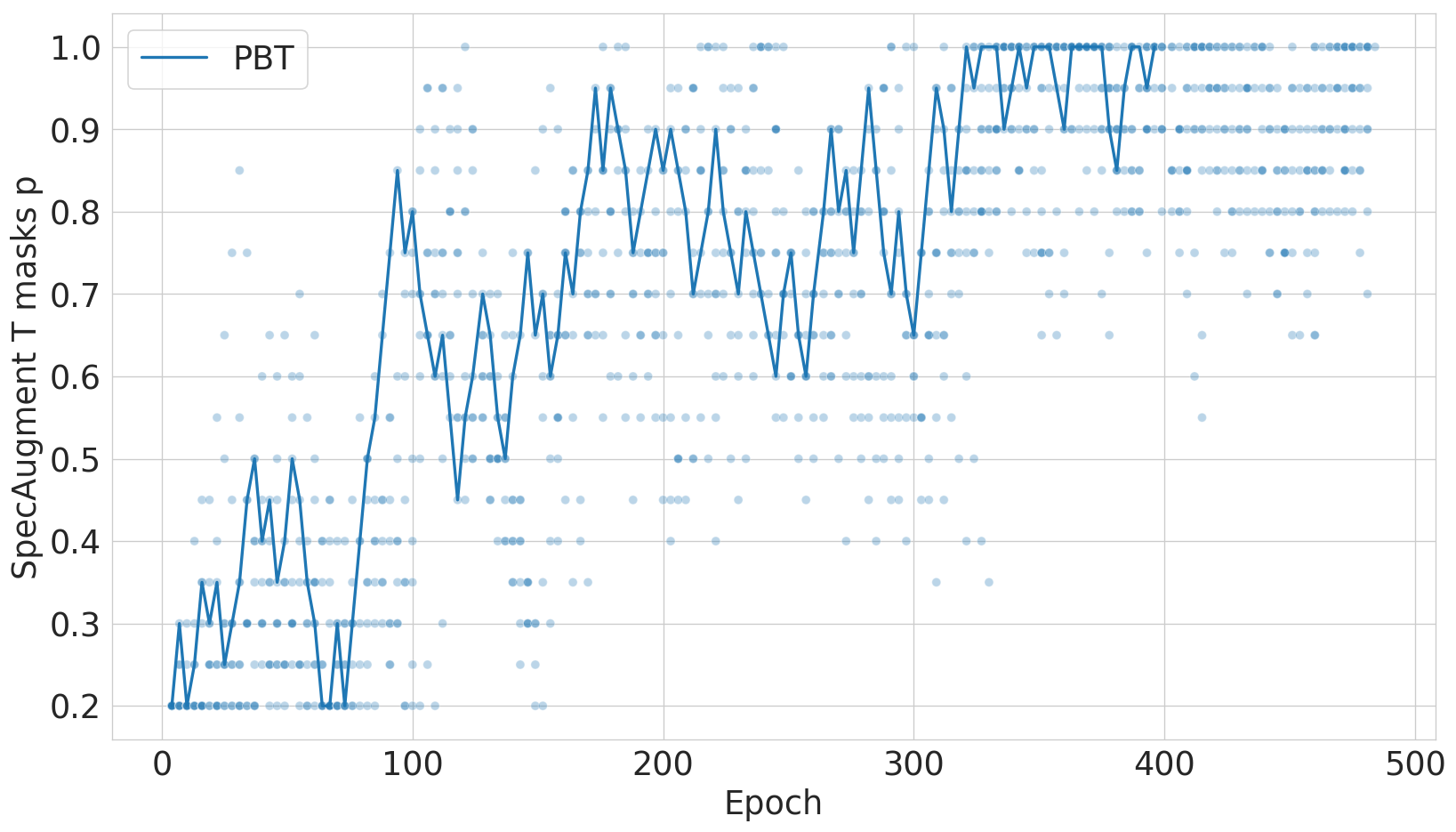}
    \includegraphics[width=0.45\textwidth]{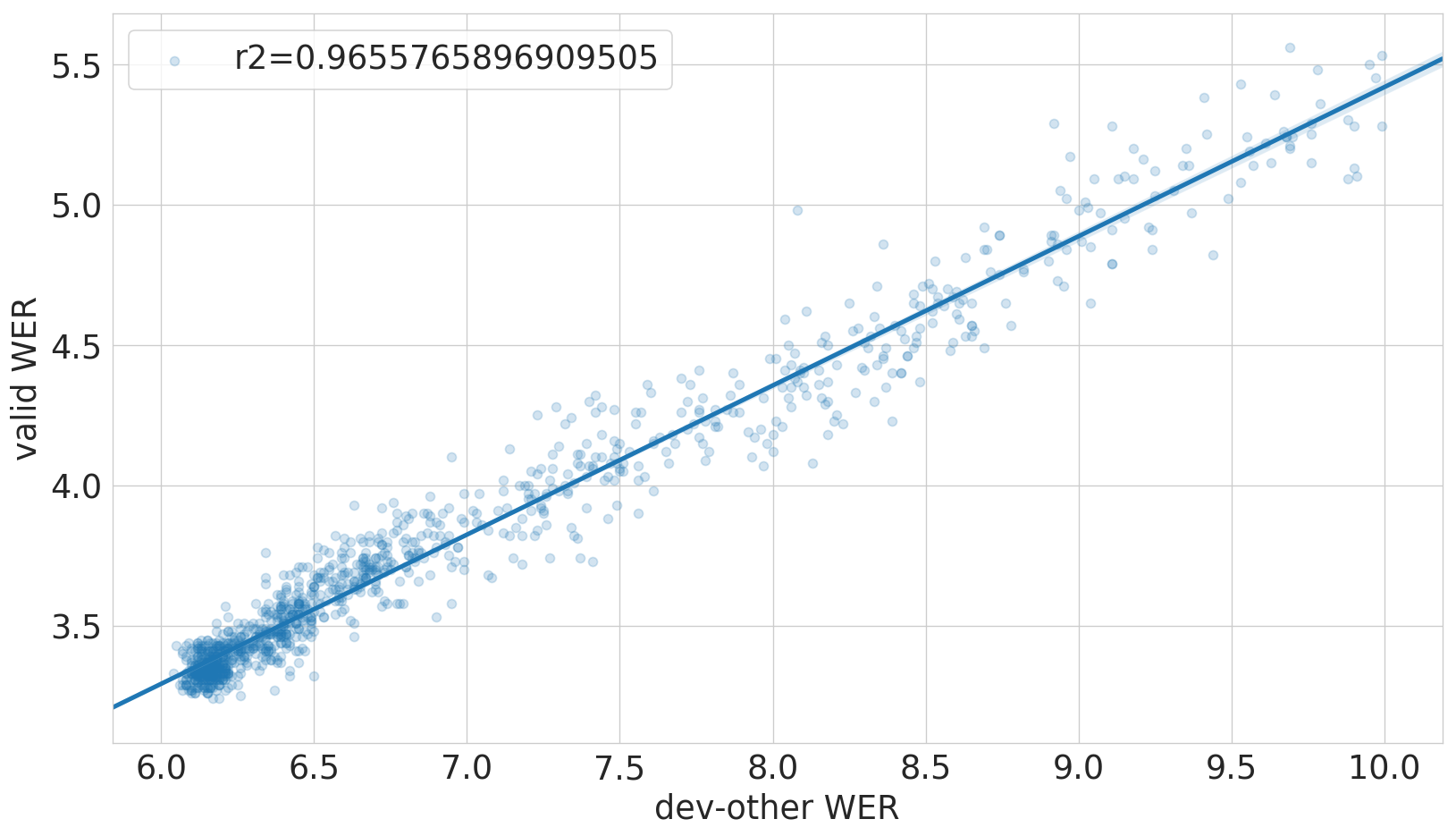}
    \subcaption{Schedule for best PBT model on \devother{} (line) and population values (dots) for SpecAugment's T masking parameters T, N and P. For comparison, commonly used values are T=100, N=2, P=1.0}\label{fig:1b}
\caption{PBT achieves superior final performance by discovering schedules of hyperparameters, while training in a similar wall clock time}
\label{w2l_devother}
\end{figure}

\begin{figure}
\centering
\includegraphics[width=0.45\textwidth]{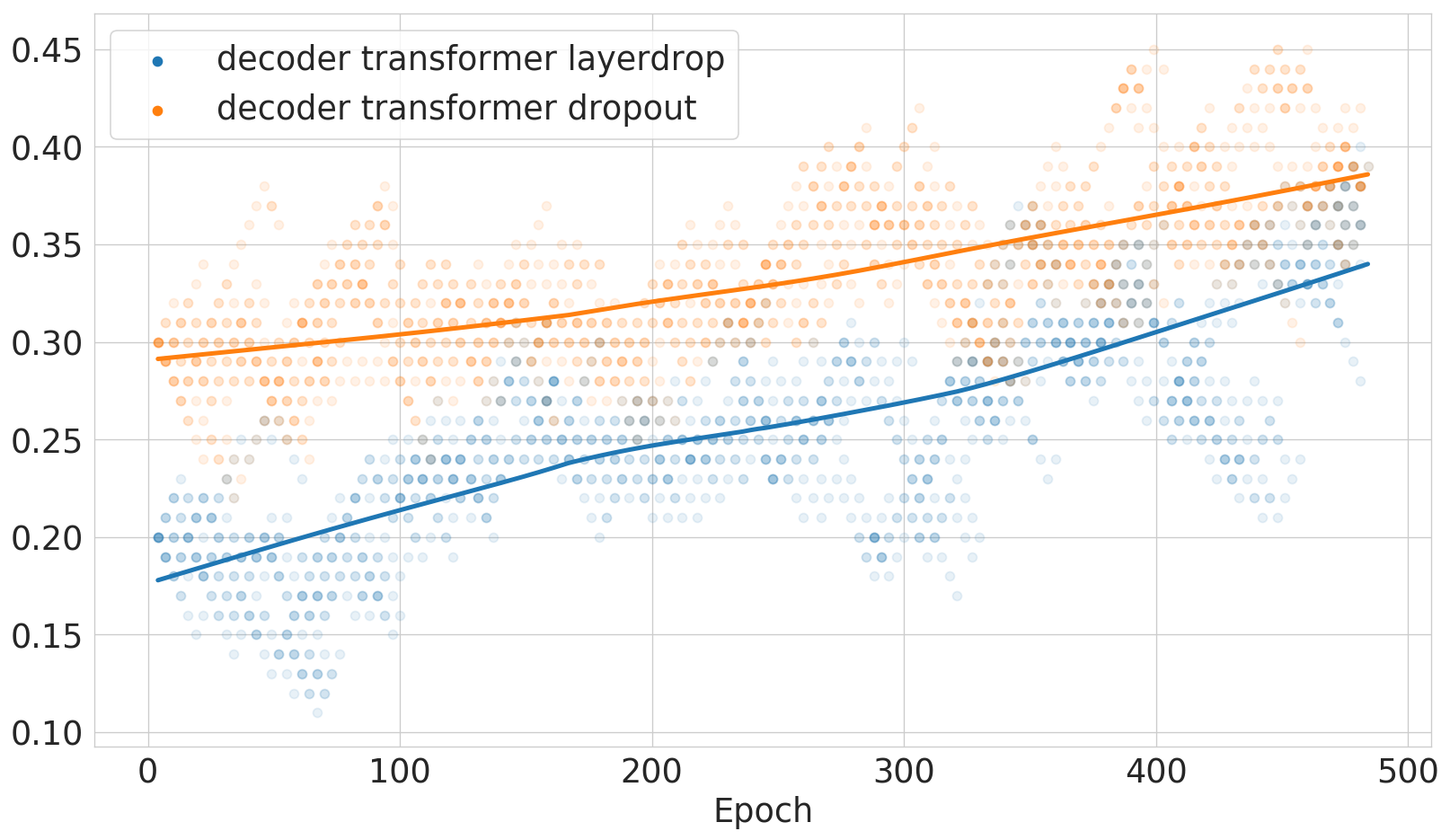}
\includegraphics[width=0.45\textwidth]{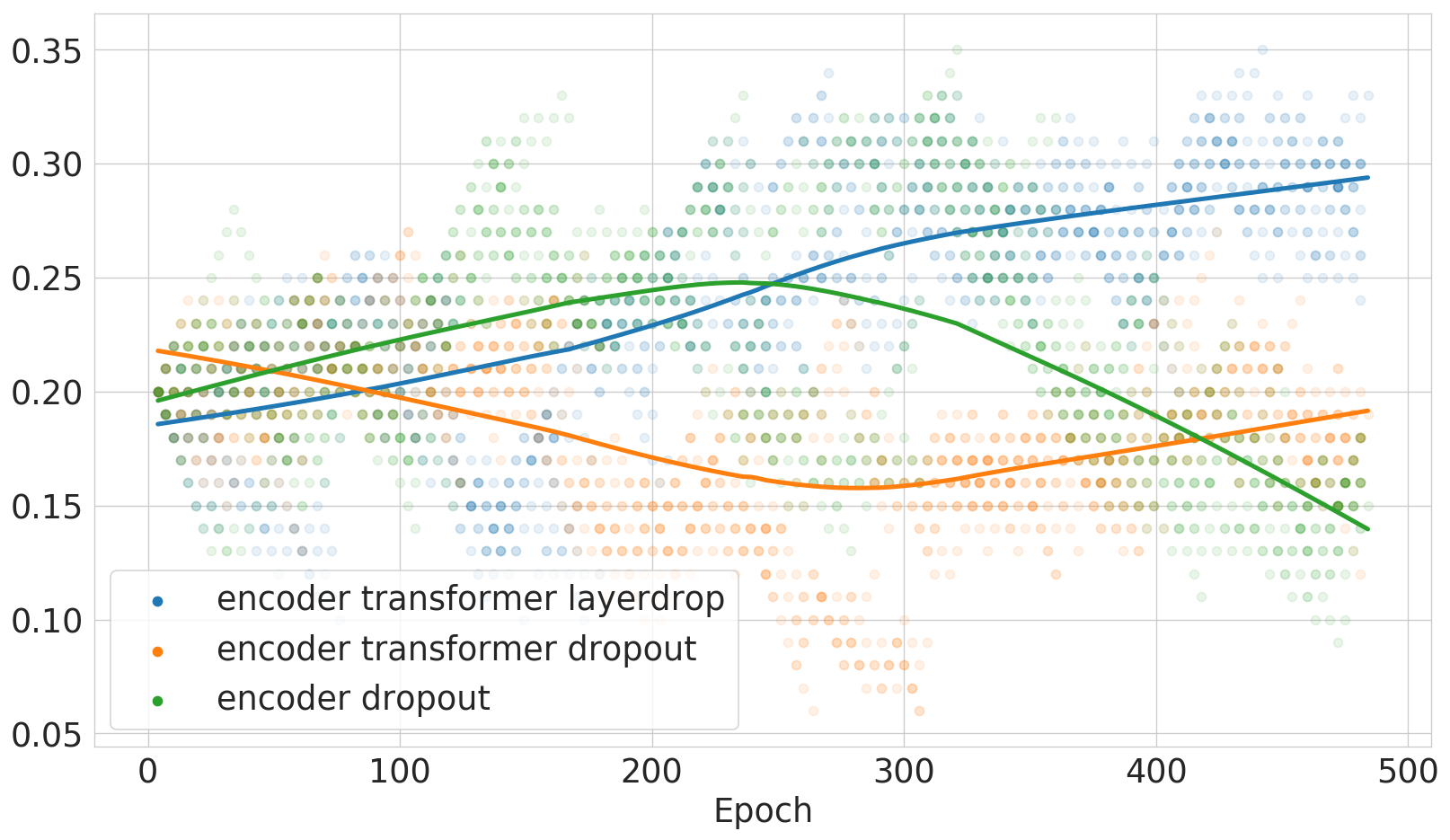}
\caption{Regularization parameters values (dots) in the population. A Lowess has been fitted to the points to show trends for each parameter.}
\label{fig:w2l_reg}
\end{figure}

After 3 days of training with PBT, we select the best performing model on \devclean{} and \devother{} in terms of Viterbi WER, then evaluate and decode them on \testclean{} and \testother{}. We report our results in Table \ref{tab:libriWER}. Note that no fine-tuning was done on those models. We reach state of the art results on \librispeech{} \testclean{}, and beat the previous state of the art for end to end models on \testother{} without additional data, by a significant margin when comparing with the same decoding setup.

\subsection{Hyperparameters schedules discovered with PBT}

After a PBT has run, it is possible to analyze how the values for different hyperparameters changed over the course of the training. One way to do that is to track the distribution of values in the current population for each epoch.
Another insight is the best parameters schedule: the values of the parameters taken by all the ancestors of the best final model. We report these distributions and schedules for SpecAugment parameters relative to the time masking in Figure~\ref{fig:1b}. Surprisingly, we observe that the values taken by PBT for SpecAugment significantly differ from those usually used in the literature. In particular, we observe that regularization with time masking is much stronger than regularization with frequency masking. While the latter varies around the values proposed in \cite{park2019specaugment}, the former is way above.

Data augmentation adds noise to the training data, potentially slowing down convergence at the beginning, but helps to avoid over-fitting, which usually happens later. The schedule discovered by PBT seems to take the best of both worlds, by gradually increasing the magnitude of time masking during the first 300 epochs.

\subsection{Importance of the schedule}
Models trained with PBT achieve a better final performance than the baseline. This could be a consequence of the specific schedule found during the training, or because of better values for hyperparameters. In Table~\ref{tab:mutation_settings}, we indeed reported that PBT models end up using radically different values for data augmentation at the end of the training, compared to the static values we used in the baseline. In Figure~\ref{fig:w2l_devother_wer}, we repeat the baseline training using those final values instead - all others hyperparameters remaining unchanged: it converges to a model that performs much worse than the PBT run or the baseline. We can conclude that the superior performance of PBT cannot be reduced to the final values found for data augmentation.

\label{sec:experiments}

\section{Discussion}
\label{sec:discussion}

PBT makes use of parallelization to discover hyperparameter schedules without wasting compute resources on unpromising models.
In theory, the search space grow exponentially with each hyperparameter. In practice many equivalent sets of hyperparameters coexist, and smooth changes of several hyperparameters results in smooth changes of the model's output. This is particularly the case for data augmentation hyperparameters.
In this work, we have shown that PBT can be used for challenging tasks on bigger models than previously demonstrated, without compromising on the training wall time, and even with relatively small populations.
Our speech recognition model has significantly superior final performance and reaches the state of the art when trained with PBT, without manual fine tuning.

In the future, we would like to see automated tuning methods discover schedules for more parameters. For example in the case of the learning rate, hand-crafted schedules have become prevalent in the literature as it has a significant impact on the convergence, yet their tuning can be cumbersome. The selection scheme we use here is not efficient in handling mutation where the short term gains are not aligned with final model fitness, which is often the case for the learning rate.

\clearpage
\newpage
\bibliographystyle{plain}

\bibliography{main}

\clearpage
\newpage
\section*{Appendix}

\subsection{PBT validation dataset}
\begin{figure}[h]
\centering
\includegraphics[width=0.45\textwidth]{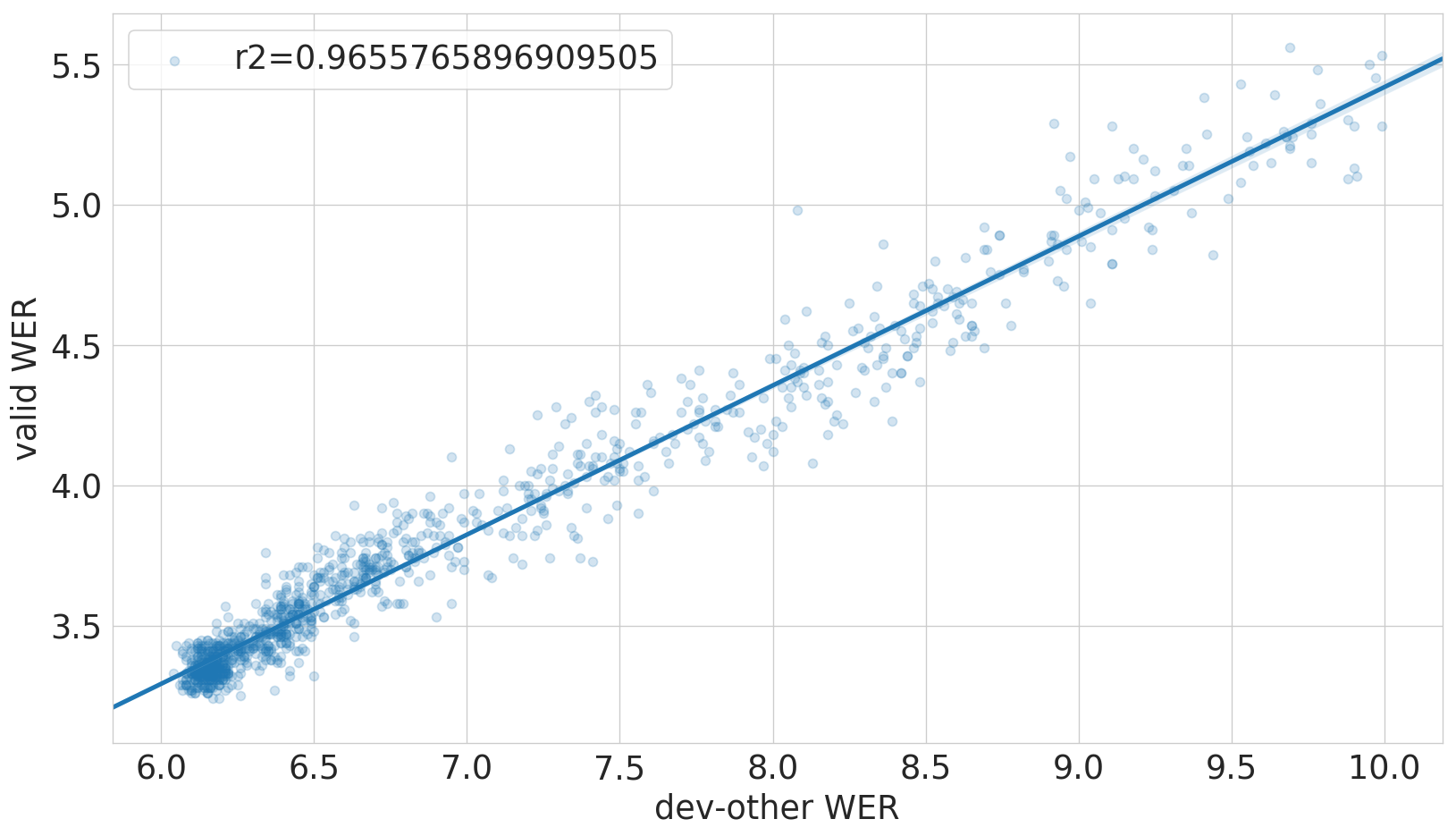}
\caption{Correlation between PBT validation loss and \devother{} for checkpoints in the PBT run. In particular, the scale is different for both datasets. This is because we extracted the PBT validation dataset from \tof, which has a different distribution from \devother{}}
\label{fig:w2l_pbt_valid_corr}
\end{figure}

\subsection{Population management details}
Given information about all previously evaluated checkpoints, workers need to find what model to train next. It is important to note that we run many workers in parallel and asynchronously: we start training models of generation $G+1$ while models of generation $G$ are still training and not evaluated yet.

\paragraph{Finding a parent model to train}
We use the initiator based evolution introduced in \cite{li2019generalized} to select a parent model. First, we compute the last completed generation $G$, as the latest generation for which we have at least 2 evaluated checkpoints.

Then, an \emph{initiator} of generation $g$ in $\{G - 2, G - 1, G \}$ is sampled from non-initiated checkpoints, and immediately marked as initiated. An \emph{opponent} checkpoint from generation $G$ or $G - 1$ is sampled at random, and plays a matchup against the \emph{initiator}. The winner of this matchup is then used as a new parent. Note that a matchup is not symmetrical for both its participants: if the initiator wins, the opponent still has an opportunity to have an offspring if it is not initiated yet. However, it the opponent wins, the initiator can only have an offspring if it is selected as an opponent later on, and wins the corresponding matchup. Also, because the choice of the opponent is stochastic, it prevents worst performing models from being systematically eliminated, helping preserve population diversity. We found that using this scheme makes trainings more robust since these models sometimes become better on the long run.

\paragraph{Matchup winner} The procedure to determine the winner between two checkpoints is critical, as it determines the trade off between exploration and exploitation. As noted above, this function is not symmetrical: it is biased toward the initiator to preserve some diversity in the population. In addition, because this procedure has to be agnostic of the loss value or scaling, we only compare the loss rank percentile in the last 2 generations. For a checkpoint of generation $g$, the rank percentile ranges between 0 (best in generations $\{g - 1, g\}$) to 1 (worst in generations $\{g - 1, g\}$), 0.5 is the median loss. Algorithm~\ref{listing:matchup} summarizes this matchup process.

\begin{lstlisting}[language=python,caption={Matchup procedure},label={listing:matchup}]
# given 2 checkpoints, returns the winner
def matchup_winner(init, opp):
    pct_init = rank_pctile(init)
    pct_opp = rank_pctile(opp)
    if pct_init - 0.25 < pct_opp:
        return init
    return opp
\end{lstlisting}

\end{document}